\documentclass[11pt,a4paper]{article}
\usepackage[hyperref]{acl2017}
\usepackage{times}
\usepackage{latexsym}
\usepackage{booktabs}
\usepackage{url}

\usepackage{graphicx}
\usepackage{color}
\usepackage{mathtools, cuted}

\aclfinalcopy 
\def\aclpaperid{8} 

\title{Automatic Summarization of Online Debates}

\author{Nattapong Sanchan, Ahmet Aker \and Kalina Bontcheva \\\\ Natural Language Processing Group, Department of Computer Science, \\ The University of Sheffield, 211 Portobello, Sheffield, United Kingdom\\ \{nsanchan1, ahmet.aker, k.bontcheva\}@sheffield.ac.uk \\ \url{https://www.sheffield.ac.uk/dcs}}

\date{}

\begin{document}
\maketitle

%
%
%
\begin{abstract}
Debate summarization is one of the novel and challenging research areas in automatic text summarization which has been largely unexplored. In this paper, we develop a debate summarization pipeline to summarize key topics which are discussed or argued in the two opposing sides of online debates. We view that the generation of debate summaries can be achieved by clustering, cluster labeling, and visualization. In our work, we investigate two different clustering approaches for the generation of the summaries. In the first approach, we generate the summaries by applying purely term-based clustering and cluster labeling. The second approach makes use of X-means for clustering and Mutual Information for labeling the clusters. Both approaches are driven by ontologies.  We visualize the results using bar charts. We think that our results are a smooth entry for users aiming to receive the first impression about what is discussed within a debate topic containing waste number of argumentations.  
\end{abstract}

%
%
%
\section{Introduction}
As the number of Internet users has been growing significantly, information is published and stored digitally in textual forms. Online debate is one example of the information which has been massively published. As more and more debate content increases, it becomes a difficult task to easily and quickly discover key arguments that are expressed in the vast amount of debate data. Automatic Text Summarization can help users to extract or summarize those key arguments more efficiently and reduce the reading time.

Online debate forums normally contain two sides of debates: proponent and opponent. This structure gives opportunities for users to choose a stance (side) for a debate topic, expresses their opinions to support their propositions, and opposes other propositions. In this paper, we explore online debates which are related to the existence of global warming. A side of proponents believes in the existence of global warming and the other side, the opponents, says that global warming is not true. When the proponents and the opponents express their sentiments, opinions, and pieces of evidence to support their propositions, arguments between them arise. 


In this paper we aim to summarize online debates about global warming. In our approach we first extract salient sentences from the two opposing sides of debates (i.e. arguments published by users). Next, we cluster them by some sort of similarity. For clustering we investigate two different approaches. Our first approach is a term-based clustering approach. The second approach is based on flat clustering, namely X-means, which can automatically determine the number of clusters. Ontologies are used as the backbone for both approaches. Ontologies have been used broadly in automatic text summarization studies. However, to the best of our knowledge, this approach has not yet been applied for summarizing online debates, especially when our purpose is to capture arguments conversed in both opposing sides. Once clusters are generated, labels representing the clusters are extracted. Again we follow two approaches. The first approach is a simple one and selects as a label an ontological term that is shared by all salient sentences within the cluster. The second labeling approach extracts such a term, based on Mutual Information (MI). The resulting clusters along with their labels are visualized using bar charts. Our results show that clustering with X-means and label generation using MI is a better choice for the purpose of online debates summarization.



The rest of the paper is organized as follows. Section \ref{related_work} discusses about related work in online debate summarization. The online debate data related to the existence of global warming are elaborated in Section \ref{data}. Section \ref{framework} illustrates the system structure for developing our debate summarization system. Within the same section we also present our evaluation results.  We conclude in \ref{conclusion}.

\section{Related Work} \label{related_work}
Debate summarization is one of the novel research areas in automatic text summarization which has been largely unexplored \cite{RanadeSummarizeDebate:2013}. Examples of related work in debate summarization includes \emph{Contrastive Summarization}, \emph{Comparative Summarization}, and \emph{Debate Stance Recognition}. Contrastive Summarization is the study of generating summary for two entities and finding the difference in sentiments among them \cite{LermanContrastive}. This kind of summarization requires the classification of polarity in order to \emph{contrast} opinions expressed in different sentiments \cite{CamprComparativeLSA,paul2010summarizing}. \citet{KimZ09} summarized contrastive pairs of sentences by aligning positive and negative opinions on the same aspect. In this work, contrastive sentence pairs were constructed based on two criteria: 1) choose sentences that represent a major sentiment orientation; and 2) the two sentences should have opposite opinions on the same aspect. Similarity functions were used for determining contrastive sentence pairs. Then sentence pairs were used as input for generating the final summary. The summary was aimed to help readers compare the pros and cons of mixed opinions.

Comparative Summarization aims to find the difference between two comparable entities so that sentiment classification may not be required \citep{CamprComparativeLSA}. \citet{ZhaiCpara} worked on comparative text mining problem which aimed to discover common topics in news articles and laptop reviews and to summarize commonalities and differences in a given set of comparable text collections. A probabilistic mixture model was proposed. It generates clusters of topics across all collections and in each collection of document. The model generates \emph{k} collections of specific topics for each collection and \emph{k} common topics across all collections. Each topic is characterized by multinomial word distribution (also called a unigram language model). High probability words were used as representatives of each cluster and are also included in the summary.

Debate Stance Recognition aims to detect stance of an opinion's holder in text. \citet{Somasundaran:2009:RSO} noticed that in online debate posts, people debate issues, express their favorites, oppose other stances, and argue why their thoughts are correct.  To determine positive sentiment about one target, expressing negative sentiment about the other side is a key target. For instance, in a debate ``\emph{Which mobile phone is better: iPhone VS Blackberry?}'', people supporting iPhone may give reasons to affirm why iPhone is better. In addition, they also express why Blackberry is not. On the Blackberry side, people may also find reasons to support their opinions and argue why the phone is unfavorable. Therefore, to identify stance, it is important to not only consider positive and negative sentiment, but also consider which target an opinion refers to. 

Unlike these, the study of \citet{RanadeSummarizeDebate:2013} directly tackled debate summarization problem and it is the one which is closest to our work. In that work, system summaries are extracted by ranking the smallest units of debates, called Dialogue Acts (DAs). The ranking of sentences is based on features including, words in DAs that is co-occurring in debate topic, topics with opinions expressed on it, sentence position, and sentence length features. However, this work does not explicitly highlight what is the key content to be summarized and how the debate summary is presented. This is different to our work.  On the other hand, in our work, we highlight the summarization of key content in debates and visualize them to be easily accessed by users.


%
%
%
\section{Online Debate Data} \label{data}
In our earlier work we created freely available debate dataset on climate change domain\footnote{This dataset can be downloaded at \url{https://goo.gl/3aicDN}.}, also referred as Salient Sentence Selection Dataset (SSSD). Each debate consists of two opposing sides, \emph{Agree} and \emph{Disagree}. Whereas the opinion on the \emph{Agree} side believes that global warming exists, the \emph{Disagree} side opposes this opinion. In this dataset, each debate comment was manually annotated by 5 judgments. The aim of the annotation was to select from each comment 20\% sentences that were salient and worth for inclusion in a summary.  For instance, for a comment containing 10 sentences 2 sentences were extracted by each annotator. The dataset contains 11 debate topics with 341 comments in total. Average number of comments for a topic is 31 comments, with the minimum and maximum of 5 and 103 comments respectively.

%
%
%

\begin{figure*}
\centering
{\includegraphics[width=15.5cm]{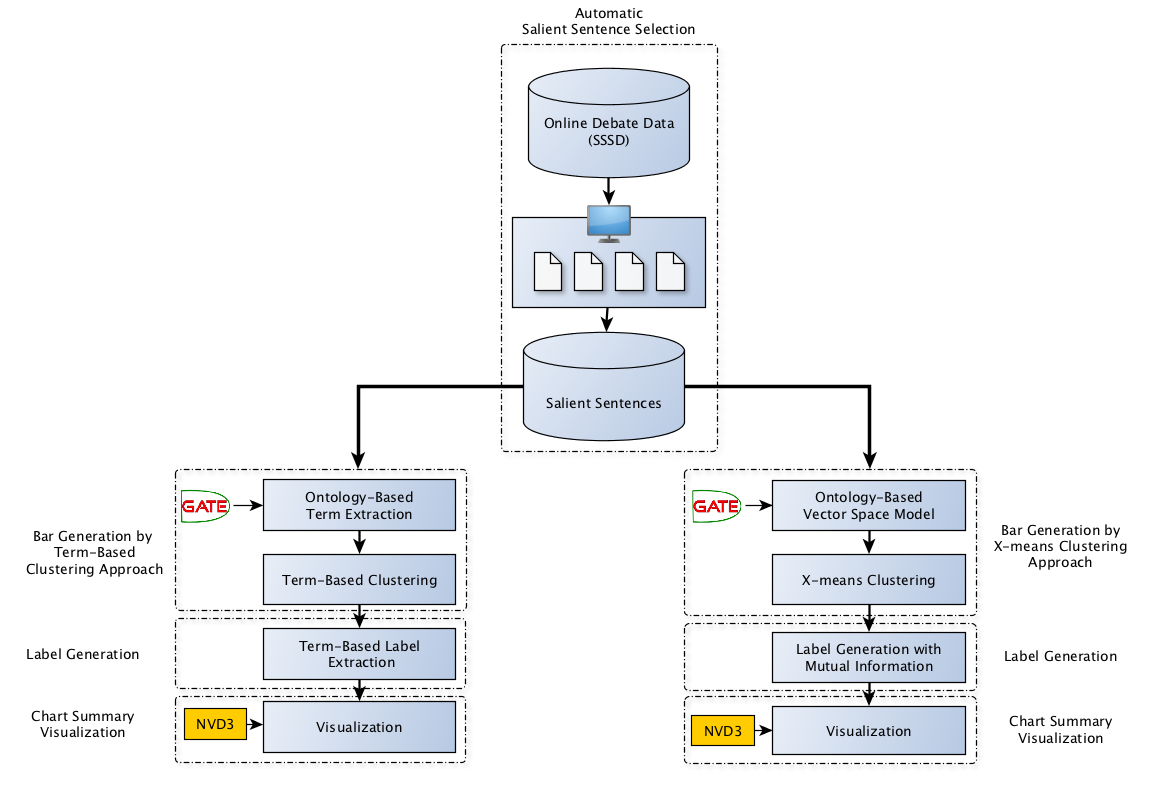}\label{fig_framework} }
\caption {The framework for generating the Chart Summary for online debate summarization} 
\end{figure*}

\section{Framework} 
\label{framework}

To generate a bar chart representing a summary of an online debate topic we proposed a pipeline with two branches where each branch presents either a term-based clustering and the term-based labeling method or X-means based clustering and the MI labeling strategy. The flow of the pipeline is shown in Figure 1. The system assumes an input of \textit{n} comments from the agree and disagree sides. Each comment consists of several sentences. We aim to select the most salient sentences from each comment, cluster the salient sentences according to their content similarity, generate clusters representing labels, and finally visualize the results using a bar chart summary. In the following sections we introduce each of the components and provide evaluation results.

\subsection{Automatic Salient Sentence Selection}

For the shown pipeline in Figure 1 we used an extractive automatic summarization system reported in our earlier work.\footnote{\url{https://goo.gl/xqVeJf}.} There are 8 main features defined in this system. Those features include sentence position (SP), sentence length (SL), title words (TT), the presence of conjunctive adverbs (CJ), cosine similarity of topic signatures\footnote{We used an approach described by \cite{Lin:2000:AAT:990820.990892} to obtain a list of topic signatures. We extract the topic signatures from our dataset which is related to climate change.} and sentences (COS\_TPS), cosine similarity of climate change terms\footnote{The terms are obtained by aggregating document keywords from online news media coverage on climate change.} and sentences (COS\_CCTS), cosine similarity of sentence and title words (COS\_TTS), and the semantic similarity of sentence and title words (COS\_STT) using Doc2Vec. Additionally, we also investigated the Combination of features (CB) in the salient sentence selection. For a given comment the system extracts 20\% sentences from it that are deemed as salient.


\subsubsection{Evaluation}

We used ROUGE-1, ROUGE-2, and ROUGE-SU4 evaluation metrics to evaluate the quality of the system summaries, i.e. the selection of salient sentences. As reference summaries we used the manually generated summaries from the freely available dataset, SSSD. Our results revealed that sentence position outperforms other features indicating that the most salient sentences are always. In addition, other useful key features are debate title words feature, and cosine similarity of debate title words and sentences feature. The complete set of results are shown in Table \ref{table_rouge_scores}.

\begin{table*}[ht]
\centering
\scalebox{0.85}{
\begin{tabular}{|c|c|c|c|c|c|c|c|c|c|}
\hline

\textbf{ROUGE-N} & \textbf{CB} & \textbf{CJ} & \textbf{COS\_CCT} & \textbf{COS\_TTS} & \textbf{COS\_TPS} & \textbf{SL} & \textbf{SP} & \textbf{COS\_STT} & \textbf{TT} \\ \hline
\textbf{R-1} & 0.4773 & 0.4988 & 0.3389 & 0.5630 & 0.3907 & 0.4307 & \textbf{0.6124} & 0.4304 & 0.5407 \\ \hline
\textbf{R-2} & 0.3981 & 0.4346 & 0.2558 & 0.5076 & 0.2986 & 0.3550 & \textbf{0.5375} & 0.3561 & 0.4693 \\ \hline
\textbf{R-SU4} & 0.3783 & 0.4147 & 0.2340 & 0.4780 & 0.2699 & 0.3335 & \textbf{0.4871} & 0.3340 & 0.4303 \\ \hline
\end{tabular}}
\caption{ROUGE scores derived from each feature in the Salient Sentence Selection task}\label{table_rouge_scores}
\end{table*}

\subsection{Term-Based Clustering} 
\label{sec:termBasedClustering}

To perform clustering we used terms extracted through ontologies. We employed the English ClimaPinion service\footnote{http://services.gate.ac.uk/decarbonet/sentiment/} from the DecarboNet project\footnote{https://www.decarbonet.eu} as the background knowledge to capture climate change topics and extract from each salient sentence topical terms. To obtain clusters we grouped sentences containing the same label within the same cluster. If a sentence contained more than one term then it was assigned to several groups allowing the sentence to be soft-clustered.\footnote{Within a cluster all sentences must share one particular term but each sentence may contain other terms that are not shared by other sentences within the same cluster.} Also note, terms with the same semantic meaning can be expressed differently. To address this, for each label, we obtained a list of its synonyms from WordNet \citep{miller1995wordnet}. If the labels shared common synonyms, we considered they are the same labels. Consequentially, the sentences automatically annotated with such labels were merged to the same clusters. 


\subsubsection{Evaluation}

The evaluation of the ontology based term extraction has been evaluated somewhere else. By consisting of two environmental ontologies, GEMET (GEneral Multilingual Environmental Thesaurus) and Reegle, the ClimaPinion yields great results in recognizing environmental terms in text, with the precision, recall, and F1 measure of 85.87\%, 53.05\%, and 65.58\% respectively \citep{gateServices}.

\begin{table}[ht]
\centering
\caption{Average silhouette scores derived from the term-based clustering approach}
\label{avg_silhouetteOntology}
\scalebox{0.85}{
\begin{tabular}{@{}cc@{}}
\toprule
\textbf{Number of Clusters} & \textbf{Average Silhouette Score} \\ \midrule
39                          & 0.0000                            \\ \bottomrule
\end{tabular}}
\end{table}

The results derived from the term-based clustering approach are evaluated with the Silhouette index \citep{Rousseeuw:1987}. Silhouette evaluates the clustering performance by determining the appropriateness of documents assigned to a cluster rather than the other clusters. These documents are represented as coordinates. Silhouette calculates the pairwise difference in both inter-cluster and intra-cluster distance. We calculated an average silhouette score and reported it in Table \ref{avg_silhouetteOntology}. As shown in the table, the system generated 39 clusters based on the climate change terms annotated by the ClimaPinion service. It achieved the silhouette score around zero, similar to the work presented by \citet{Wang:2017}. The interpretation based on the score is the data points are assigned nearly to the decision boundaries of the clusters. Especially, when salient sentences contain multiple climate change topics, clear clustering boundaries are difficult to achieve. This circumstance indicates that such a simple clustering approach is less applicable for grouping semantically similar sentences together and that the task asks for more sophisticated ways for achieving a better performance. We will discuss an alternative solution in Section \ref{sec:x-meansClustering}.

\subsection{Term-Based Label Extraction} 
\label{sec:termBasedLabeling}

After grouping salient sentences together, the groups or clusters should be given labels which clearly reflect the content in the clusters \cite{AkerPKFBHG16}. Similar to the clustering approach, where we grouped salient sentences by the ontological term they share, we used the sharing term as the label to represent the cluster. This is based on the assumption that the climate change terms which are annotated in the sentences do already elaborate the central meaning of the clusters. 

\subsubsection{Evaluation} \label{eval_termbase_labels}

In the labeling evaluation, we compared the system labels against the baseline labels. We generated the baseline labels by applying \emph{tf*idf}. It is a common approach in most information retrieval systems \citep{Ponte} which consists of two main components, \emph{tf} and \emph{idf}. In our experiment, \emph{tf} indicates the frequency of terms occurs in a cluster\footnote{Since sentences can carry more than a term it is likely that a cluster has several climate change terms.}. \emph{idf} presents the number clusters in which the term occurs. These components allow us to reduce common terms in the clusters and discover more discriminative terms having fairly low term frequency in the clusters. To determine the candidate labels, we calculated the score for each term by the multiplication of \emph{tf} and \emph{idf}. The term with the top score was chosen as the cluster label. 

\begin{figure}[ht]
\centering
{\includegraphics[width=7.5cm]{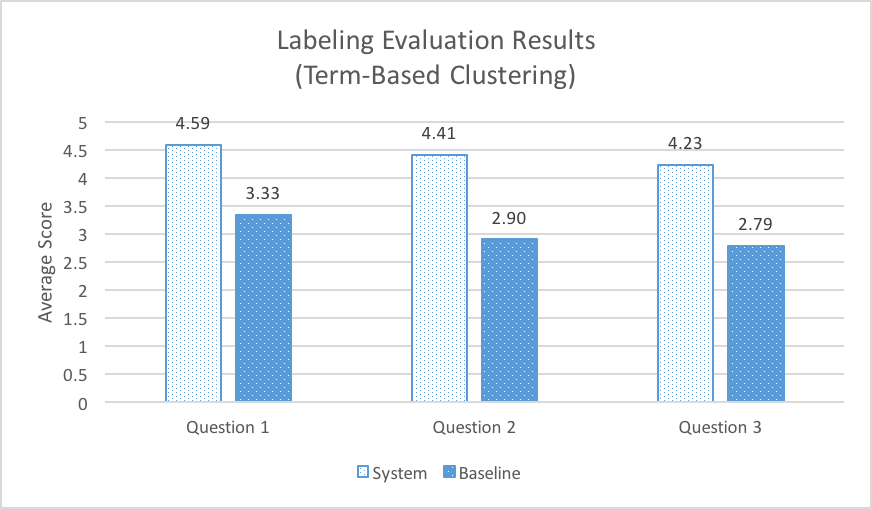}}
\caption {The labeling evaluation performed on the term-based clustering approach. The average preference scores generated by 3 participants on a scale 1: strongly disagree to 5: strongly agree}
\label{fig_results_label_termbase}
\end{figure}

In the evaluation of cluster labels, we followed the manual evaluation method presented by \citet{AkerPKFBHG16}. We invited three participants, two PhD candidates and one researcher having background in Computer Science, to evaluate the labels. The evaluation was presented as an online form. The participants were asked to read the sentences in the given clusters and score the labels. The baseline and system labels were shown in random order. For each label, the participants were asked to answer five-point Likert scale questions, ranking from \emph{strongly disagree} (1) to \emph{strongly agree} (5). The questions include i) Question 1: By reading the label, I can understand it, ii) Question 2: This label is a complete phrase, and iii) Question 3: This label precisely reflects the content of the sentences in cluster. Along with the three questions, we presented 13 clusters with a maximum of 10 salient sentences (so that the participants are able to read the content prior to the labeling evaluation) and a minimum of 2 salient sentences. Figure \ref{fig_results_label_termbase} illustrates the results of the labeling evaluation.

As we can see from the figure, in overall, the quality of system labels outperforms the baseline labels. In Q1, the system labels compared to the baseline labels are more understandable with the average score of 4.59 and 3.33 respectively. Likewise, in Q2, the system labels are more completed phrases than the baseline with the mean difference of 1.51. Lastly, with the average preference scores of 4.23 in Q3, the system labels are more reflecting the quality of content in the clusters than those generated by the baseline having the score of 2.79. Additionally, the quality of the system labels is further confirmed by a statistical significance analysis with Mann-Whitney U Test. The test reveals that significance difference is found in the system labels ($Md_{Q1-Q3} = 5,\; n_{Q1-Q3} = 39)$ and the baseline labels ($Md_{Q1}=4, \;Md_{Q2} = 3, \;Md_{Q3} = 2),\; U_{Q1} = 363,\; U_{Q2} = 343,\; U_{Q3} = 386,\; z_{Q1} = -4.25,\; z_{Q2} = -4.36,\; z_{Q3}= -3.92, \;p < 0.01,\; r_{Q1} = 0.48,\; \;r_{Q2} = 0.49,\; r_{Q3} = 0.44$. We also measured the inter-annotator agreement using Krippendorff’s alpha coefficient\footnote{The measurement is performed using \emph{nltk metrics}, http://www.nltk.org/api/nltk.metrics.html.}. The agreement in Q1, Q2, and Q3 are 0.31, 0.27, and 0.35 respectively.

%
%
%

\subsection{X-means Clustering}
\label{sec:x-meansClustering}

In Section \ref{sec:termBasedClustering} we have shown that the idea of performing clustering based on shared terms results in poor clustering performance. The approach leads to too many clusters which are very close to each other. In this section we aim to overcome the problem of poor performance of the term-based clustering approach and use X-means \citep{xmeans}, an extended version of K-means, to cluster the salient sentences selected by the summarization system. One of the benefits of X-means is that it is able to automatically detect the number of clusters. By computing the Bayesian Information Criterion (BIC) scores, X-means decides if cluster centroids should be split. We applied ontology-based vector space model approach to create vectors as the similarity inputs for X-means. 

\subsubsection{Similarity Measurement}

To enable X-means to process the clustering, a similarity needs to be defined to determine which sentences are close to each other. In the definition of our similarity measurement, the automatic selected salient sentences are transformed into vectors using the Vector Space Model (VSM). In the document indexing stage, we employed the ontologies to automatically annotate key climate change terms in the SSSD. The employment of ontology-based approach benefits the transformation of words to vectors by help capturing relevance of specific topics. We derived 64 significant climate change topics. Term frequency was counted for each term to generate vectors for each sentence. To generate a similarity matrix, cosine similarity measure was used to calculate cosine similarity scores among the vectors. After the similarity matrix was constructed, we applied a Principal Component Analysis (PCA)\footnote{sklearn.decomposition.PCA: https://goo.gl/QqiWec} for the dimensionality reduction.

\subsubsection{Evaluation}

Similar to the ontology term-based clustering we evaluated the results of the X-means clustering using Silhouette index. Results are reported in Table \ref{avg_silhouette}. 

As shown in the table, the average silhouette score is derived from the calculation based on the similarity definition obtained by the ontology-based vector space model. We achieved a high silhouette score of 0.9878, with the total number of 19 generated clusters. A silhouette close to 1.0 indicates good cohesion and separation of the clustering results, meaning that the average distance from a coordinate in a cluster to the other coordinates within its own cluster is less than the average distance to all coordinate in the nearest cluster. In addition, when the score is close to 0, the coordinates in the clusters are nearly close or on the decision boundary between two neighbouring clusters. A negative silhouette score is obtained when coordinates might be assigned to wrong clusters. In other words, the coordinates are very close to the neighbouring cluster rather than the coordinates in their own clusters \citep{Rousseeuw:1987}. In our experiment, we concluded that the clustering results obtained by X-means clustering algorithm have strong clustering structure and is more appropriate for the task of summarizing debate data. 

\begin{table}[ht]
\centering
\caption{Average silhouette scores derived from X-means clustering results}
\label{avg_silhouette}
\scalebox{0.85}{
\begin{tabular}{@{}cc@{}}
\toprule
\textbf{Number of Clusters} & \textbf{Average Silhouette Score} \\ \midrule
19                          & 0.9878                            \\ \bottomrule
\end{tabular}}
\end{table}

\subsection{Label Generation with Mutual Information}

To generate labels from the X-means clusters we could have followed the same approach as described in Section \ref{sec:termBasedLabeling}, namely picking up a term that is shared by all or majority of the salient sentences within a cluster. We tried this however, to our surprise the performance was very low compared to what we achieved in Section \ref{sec:termBasedLabeling}. Nevertheless this helped us to draw two conclusions. First, the performance in Section \ref{sec:termBasedLabeling} is high because the labels were so selected that all salient sentences within a group shared that label. Second the size of the clusters was not big so that the label had high chance to be representative of the cluster. This pictures changed once the cluster size increased and also the salient sentences covered several different climate change terms. Because of this selecting a label was not about just simply selecting the term that appears in all or majority of the salient sentences. We used Mutual Information (MI) to make this decision for us.

MI is a prevalent feature selection approach that involves in the calculation of a utility measure A(\emph{t},\emph{c}). MI quantifies how much information that term \emph{t} contributing to the correct classification judgment on class \emph{c} \citep{manning2008}. The MI formula is shown in Equation \ref{eq_mi}, where \emph{U} is a random variable that holds the value \emph{$e_t$}. If a sentence contains term \emph{t}, the value of \emph{$e_t$} is \emph{1}. Otherwise, the \emph{$e_t$} is 0. \emph{C} is a random variable that holds the value \emph{$e_c$}. The value of \emph{$e_c$} is 1 indicating that a sentence is in class \emph{c} and it is 0 if it is not.

\newpage

\begin{strip}
\begin{equation} \label{eq_mi}
  I(U;C) = \sum_{e_t\in{\{1,0\}}} \sum_{e_c\in{\{1,0\}}} P(U = e_t, C = e_c)\log_2 \frac{P(U=e_t,C=e_c)}{P(U=e_t)P(C=e_c)^\prime}
\end{equation}

\begin{equation} \label{eq_mi2}
  I(U;C) = \frac{N_{11}}{N}\log_2\frac{NN_{11}}{N_{1.}N_{.1}} + \frac{N_{01}}{N}\log_2\frac{NN_{01}}{N_{0.}N_{.1}} + \frac{N_{10}}{N}\log_2\frac{NN_{10}}{N_{1.}N_{.0}} + \frac{N_{00}}{N}\log_2\frac{NN_{00}}{N_{0.}N_{.0}}
\end{equation}
\end{strip}

To calculate the mutual information scores for candidate terms, we applied the maximum likelihood estimation of probability as shown in Equation \ref{eq_mi2} \citep{manning2008}. From the equation, \emph{N} refers to the counts of sentences in which its subscripts take the values of \emph{$e_t$} and \emph{$e_c$}. For instance, \emph{$N_{01}$} refers to the number of sentences that do not containing term t (\emph{$e_t = 1$}) but in class \emph{c} (\emph{$ec=1$}). \emph{$N_{1.}$} is derived from the addition of \emph{$N_{10}$} and \emph{$N_{11}$}. \emph{N} refers to the total number of sentences. In each cluster, we calculated the score of each candidate term. The term with the higher MI score was selected as the cluster label for that cluster.

\subsubsection{Labeling Evaluation}
In order to evaluate the system labels generated by the results derived from X-means clustering approach, we applied the same evaluation procedure as well the baseline discussed in Section \ref{eval_termbase_labels}. The results are illustrated in Figure \ref{fig_results_label_termbase}. As can be seen from the figure, the average preference scores of the system outperform the baseline. In Q1, the system labels are more understandable than the baseline, with the mean difference of 0.10. In Q2, the system labels more completed phrases than the baseline labels, with a higher mean score of 0.13. Lastly, in Q3, the system labels are still better than the baseline labels with the mean difference of 0.05. The system labels are more meaningful for presenting the central meaning of the content in the clusters. However, as there is a slight difference between the results of the system labels and baseline labels, Mann–-Whitney U test reveals that no significant difference, with the z values of -0.705,-0.427, and -0.389, with the significance levels of p= 0.481, 0.670, and 0.697 respectively. The values of Krippendorff's alpha, by another three participants, for Q1, Q2, and Q3 are 0.33, 0.44, and 0.56 respectively.

\begin{figure}[ht]
\centering
{\includegraphics[width=7.5cm]{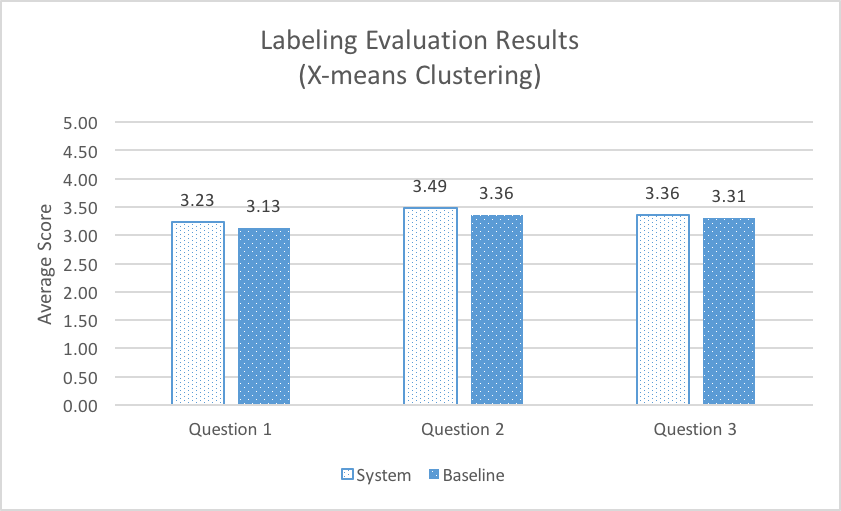}}
\caption {The labeling evaluation performed on the X-means clustering approach. The average preference scores generated by 3 participants on a scale 1: strongly disagree to 5: strongly agree}
\label{fig_results_label_xmeans}
\end{figure}


\subsection{Visualization}

\begin{figure}[ht]
\centering
{\includegraphics[width=8cm]{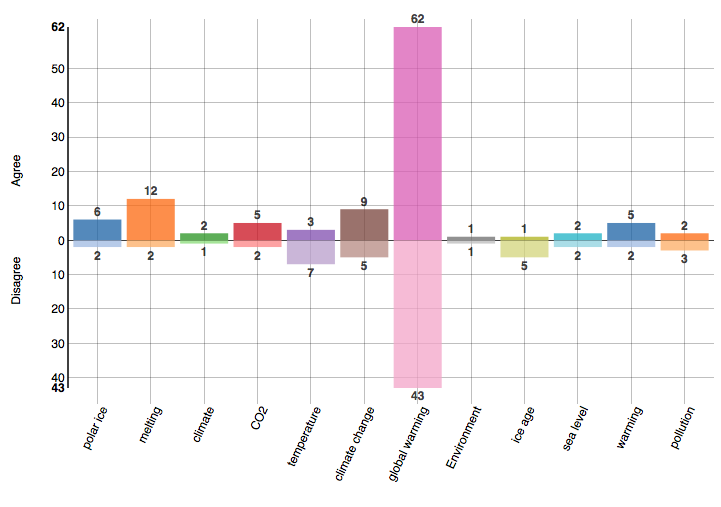}}
\caption {Chart Summary for Debate Data}
\label{fig_process02_chart_summary}
\end{figure}

\citet{SanchanBA16} have manually investigated various representation models for displaying or visualizing summaries of online debates. Unlike traditional summaries, the debates extracts have to capture main concepts discussed in both sides and enable the reader to look at those concepts from both proponent and opponent sites. The authors proposed the Chart Summary which presents the clusters by bars. Each bar is marked with the cluster label. In this work we adopt the Chart Summary for visualization purposes. An example Chart Summary is shown in Figure \ref{fig_process02_chart_summary}.



In the generation of the bars in Chart Summary, the bars are the clusters that express related content in both opposing sides. Therefore, it is important to match clusters from the two opposing sides which express the related content. We call this approach as \emph{alignment}. From the two opposing sides, we align the clusters based on the cluster labels. The clusters sharing mutual labels are aligned. For alignment, we used cosine similarity over vector spaces representing the labels. The vector also contains semantically related words enriched from WordNet. Clusters which have no pair will not be aligned and thus will not be presented in the Chart Summary. Once the pairs of aligned clusters are derived, we count the number of salient sentences in those clusters, separately in each opposing side. Those numbers represent the frequencies of the bars.

After all components of a Chart Summary are completely generated, they are exported to NVD3 JAVA script\footnote{http://nvd3.org} for the purpose of visualizing the Chart Summary. Figure \ref{fig_process02_chart_summary} illustrates a Chart Summary for the online debates data which runs on a web browser\footnote{A full version of Chart Summary can be accessed via https://goo.gl/wjBh7V.}.

\section{Conclusion} \label{conclusion}
In this paper, we investigated one of the novel and challenging problems in automatic text summarization of online debates and proposed a framework to tackle this problem. We aimed to generate Chart Summaries which represent the high-level topics of online debates. The Chart Summary is composed of three main components, including the bars, labels, and frequencies of the bars. We proposed an ontological term-based driven clustering and cluster labeling pipeline to guide the debate summary generation. In our approach we used an online service to automatically annotate climate change terms in salient sentences and to group related salient sentences into the same cluster. For clustering we investigated two variants both making use of ontological terms. The first, a simply approach, groups salient sentences by shared terms. The second approach applies X-means clustering. The evaluation has shown that the X-means approach is a better choice for clustering. We create labels to represent each cluster. Again here we investigated two different approaches both making use of ontological terms. The first approach, again a simply one, labels each cluster with the term shared by all members within the cluster. The second approach picks up the best term according to Mutual Information. The manual evaluation showed that the simple approach achieves higher results than the MI one. However, as discussed the simply approach achieved high results because of the size of the clusters and led to poor results when the size of the cluster grew which is the case with the X-means clustering. Once the clusters and labels are generated with align the pro and contra parts and visualize the results using NVD3.

In future work we plan to enrich the Chart Summary with additional details such as enabling the users to see example debates for each pair of clusters.  



\bibliography{acl2017V003}

\begin{thebibliography}{}
\expandafter\ifx\csname natexlab\endcsname\relax\def\natexlab#1{#1}\fi

\bibitem[{Aker et~al.(2016)Aker, Paramita, Kurtic, Funk, Barker, Hepple, and
  Gaizauskas}]{AkerPKFBHG16}
Ahmet Aker, Monica~Lestari Paramita, Emina Kurtic, Adam Funk, Emma Barker, Mark
  Hepple, and Robert~J. Gaizauskas. 2016.
\newblock \href{http://aclweb.org/anthology/W/W16/W16-6610.pdf}{Automatic label
  generation for news comment clusters}.
\newblock In {\em {INLG} 2016 - Proceedings of the Ninth International Natural
  Language Generation Conference, September 5-8, 2016, Edinburgh, {UK}\/}.
  pages 61--69.
\newblock
  \href{http://aclweb.org/anthology/W/W16/W16-6610.pdf}{http://aclweb.org/anthology/W/W16/W16-6610.pdf}.

\bibitem[{Campr and Jezek(2012)}]{CamprComparativeLSA}
Michal Campr and Karel Jezek. 2012.
\newblock
  \href{http://textmining.zcu.cz/publications/wseas-mcampr.pdf}{Comparative
  summarization via latent semantic analysis}.
\newblock In {\em Lastest Trends in Information Technology;Proceedings of the
  1st WSEAS International Conference on Information Technology and Computer
  Networks (ITCN '12), Proceedings of the 1st WSEAS International Conference on
  Cloud Computing (CLC '12), Proceedings of the 1st WSEAS International
  Conference on Programming Languages and Compilers (PRLC '12)\/}. WSEAS Press,
  Stroudsburg, PA, USA, Recent Advances in Computer Engineering Series 7, pages
  279--284.
\newblock
  \href{http://textmining.zcu.cz/publications/wseas-mcampr.pdf}{http://textmining.zcu.cz/publications/wseas-mcampr.pdf}.

\bibitem[{Kim and Zhai(2009)}]{KimZ09}
Hyun~Duk Kim and ChengXiang Zhai. 2009.
\newblock \href{http://dblp.uni-trier.de/db/conf/cikm/cikm2009.html}{Generating
  comparative summaries of contradictory opinions in text}.
\newblock In David Wai-Lok Cheung, Il-Yeol Song, Wesley~W. Chu, Xiaohua Hu, and
  Jimmy~J. Lin, editors, {\em CIKM\/}. ACM, pages 385--394.
\newblock
  \href{http://dblp.uni-trier.de/db/conf/cikm/cikm2009.html}{http://dblp.uni-trier.de/db/conf/cikm/cikm2009.html}.

\bibitem[{Lerman and McDonald(2009)}]{LermanContrastive}
Kevin Lerman and Ryan McDonald. 2009.
\newblock \href{http://dl.acm.org/citation.cfm?id=1620853.1620886}{Contrastive
  summarization: An experiment with consumer reviews}.
\newblock In {\em Proceedings of Human Language Technologies: The 2009 Annual
  Conference of the North American Chapter of the Association for Computational
  Linguistics, Companion Volume: Short Papers\/}. Association for Computational
  Linguistics, Stroudsburg, PA, USA, NAACL-Short '09, pages 113--116.
\newblock
  \href{http://dl.acm.org/citation.cfm?id=1620853.1620886}{http://dl.acm.org/citation.cfm?id=1620853.1620886}.

\bibitem[{Lin and Hovy(2000)}]{Lin:2000:AAT:990820.990892}
Chin-Yew Lin and Eduard Hovy. 2000.
\newblock \href{https://doi.org/10.3115/990820.990892}{The automated
  acquisition of topic signatures for text summarization}.
\newblock In {\em Proceedings of the 18th Conference on Computational
  Linguistics - Volume 1\/}. Association for Computational Linguistics,
  Stroudsburg, PA, USA, COLING '00, pages 495--501.
\newblock
  \href{https://doi.org/10.3115/990820.990892}{https://doi.org/10.3115/990820.990892}.

\bibitem[{Manning et~al.(2008)Manning, Raghavan, and Schütze}]{manning2008}
C.~D. Manning, P.~Raghavan, and H.~Schütze. 2008.
\newblock {\em Introduction to Information Retrieval\/}.
\newblock Cambridge University Press.
\newblock
  \href{http://www-csli.stanford.edu/~hinrich/information-retrieval-book.html}{http://www-csli.stanford.edu/~hinrich/information-retrieval-book.html}.

\bibitem[{Maynard and Bontcheva(2015)}]{gateServices}
Diana Maynard and Kalina Bontcheva. 2015.
\newblock Understanding climate change tweets: an open source toolkit for
  social media analysis.
\newblock In Volker Wohlgemuth Chris Preist Elina~Eriksson Vivian
  Kvist~Johannsen, Stefan~Jensen, editor, {\em Atlantis Press\/}. Morgan
  Kaufmann Publishers Inc., Atlantis Press, pages 242--250.

\bibitem[{Miller(1995)}]{miller1995wordnet}
George~A Miller. 1995.
\newblock Wordnet: a lexical database for english.
\newblock {\em Communications of the ACM\/} 38(11):39--41.

\bibitem[{Paul et~al.(2010)Paul, Zhai, and Girju}]{paul2010summarizing}
Michael~J Paul, ChengXiang Zhai, and Roxana Girju. 2010.
\newblock Summarizing contrastive viewpoints in opinionated text.
\newblock In {\em Proceedings of the 2010 Conference on Empirical Methods in
  Natural Language Processing\/}. Association for Computational Linguistics,
  pages 66--76.

\bibitem[{Pelleg and Moore(2000)}]{xmeans}
Dan Pelleg and Andrew~W. Moore. 2000.
\newblock \href{http://dl.acm.org/citation.cfm?id=645529.657808}{X-means:
  Extending k-means with efficient estimation of the number of clusters}.
\newblock In {\em Proceedings of the Seventeenth International Conference on
  Machine Learning\/}. Morgan Kaufmann Publishers Inc., San Francisco, CA, USA,
  ICML '00, pages 727--734.
\newblock
  \href{http://dl.acm.org/citation.cfm?id=645529.657808}{http://dl.acm.org/citation.cfm?id=645529.657808}.

\bibitem[{Ponte and Croft(1998)}]{Ponte}
Jay~M. Ponte and W.~Bruce Croft. 1998.
\newblock \href{https://doi.org/10.1145/290941.291008}{A language modeling
  approach to information retrieval}.
\newblock In {\em Proceedings of the 21st Annual International ACM SIGIR
  Conference on Research and Development in Information Retrieval\/}. ACM, New
  York, NY, USA, SIGIR '98, pages 275--281.
\newblock
  \href{https://doi.org/10.1145/290941.291008}{https://doi.org/10.1145/290941.291008}.

\bibitem[{Ranade et~al.(2013)Ranade, Gupta, Varma, and
  Mamidi}]{RanadeSummarizeDebate:2013}
Sarvesh Ranade, Jayant Gupta, Vasudeva Varma, and Radhika Mamidi. 2013.
\newblock \href{https://doi.org/10.1145/2502069.2502076}{Online debate
  summarization using topic directed sentiment analysis}.
\newblock In {\em Proceedings of the Second International Workshop on Issues of
  Sentiment Discovery and Opinion Mining\/}. ACM, New York, NY, USA, WISDOM
  '13, pages 7:1--7:6.
\newblock
  \href{https://doi.org/10.1145/2502069.2502076}{https://doi.org/10.1145/2502069.2502076}.

\bibitem[{Rousseeuw(1987)}]{Rousseeuw:1987}
Peter Rousseeuw. 1987.
\newblock \href{https://doi.org/10.1016/0377-0427(87)90125-7}{Silhouettes: A
  graphical aid to the interpretation and validation of cluster analysis}.
\newblock {\em J. Comput. Appl. Math.\/} 20(1):53--65.
\newblock
  \href{https://doi.org/10.1016/0377-0427(87)90125-7}{https://doi.org/10.1016/0377-0427(87)90125-7}.

\bibitem[{Sanchan et~al.(2016)Sanchan, Bontcheva, and Aker}]{SanchanBA16}
Nattapong Sanchan, Kalina Bontcheva, and Ahmet Aker. 2016.
\newblock \href{https://doi.org/10.17562/PB-54-10}{Understanding human
  preferences for summary designs in online debates domain}.
\newblock {\em Polibits\/} 54:79--85.
\newblock
  \href{https://doi.org/10.17562/PB-54-10}{https://doi.org/10.17562/PB-54-10}.

\bibitem[{Somasundaran and Wiebe(2009)}]{Somasundaran:2009:RSO}
Swapna Somasundaran and Janyce Wiebe. 2009.
\newblock \href{http://dl.acm.org/citation.cfm?id=1687878.1687912}{Recognizing
  stances in one debates}.
\newblock In {\em Proceedings of the Joint Conference of the 47th Annual
  Meeting of the ACL and the 4th International Joint Conference on Natural
  Language Processing of the AFNLP: Volume 1 - Volume 1\/}. Association for
  Computational Linguistics, Stroudsburg, PA, USA, ACL '09, pages 226--234.
\newblock
  \href{http://dl.acm.org/citation.cfm?id=1687878.1687912}{http://dl.acm.org/citation.cfm?id=1687878.1687912}.

\bibitem[{Wang and Koopman(2017)}]{Wang:2017}
Shenghui Wang and Rob Koopman. 2017.
\newblock \href{https://doi.org/10.1007/s11192-017-2298-x}{Clustering articles
  based on semantic similarity}.
\newblock {\em Scientometrics\/} 111(2):1017--1031.
\newblock
  \href{https://doi.org/10.1007/s11192-017-2298-x}{https://doi.org/10.1007/s11192-017-2298-x}.

\bibitem[{Zhai et~al.(2004)Zhai, Velivelli, and Yu}]{ZhaiCpara}
ChengXiang Zhai, Atulya Velivelli, and Bei Yu. 2004.
\newblock \href{https://doi.org/10.1145/1014052.1014150}{A cross-collection
  mixture model for comparative text mining}.
\newblock In {\em Proceedings of the Tenth ACM SIGKDD International Conference
  on Knowledge Discovery and Data Mining\/}. ACM, New York, NY, USA, KDD '04,
  pages 743--748.
\newblock
  \href{https://doi.org/10.1145/1014052.1014150}{https://doi.org/10.1145/1014052.1014150}.

\end{thebibliography}

\bibliographystyle{acl_natbib}

\appendix

\end{document}